\pdfoutput=1

\documentclass[11pt]{article}

\usepackage[final]{acl}

\usepackage{times}
\usepackage{array}
\usepackage{latexsym}

\usepackage[T1]{fontenc}

\usepackage[utf8]{inputenc}

\usepackage{microtype}

\usepackage{inconsolata}

\usepackage{graphicx}

\usepackage[disable]{todonotes}

\title{GerPS-Compare: Comparing NER methods for legal norm analysis}

\author{
  \textbf{Sarah T. Bachinger\textsuperscript{1,2}},
  \textbf{Christoph Unger\textsuperscript{3}},
  \textbf{Robin Erd\textsuperscript{1,2}},
  \textbf{Leila Feddoul\textsuperscript{1,2}},
\\
  \textbf{Clara Lachenmaier\textsuperscript{3}},
  \textbf{Sina Zarrieß\textsuperscript{3}},
  \textbf{Birgitta König-Ries\textsuperscript{1}},
\\
\\
  \small\textsuperscript{1}Heinz Nixdorf Chair for Distributed Information Systems, Friedrich Schiller University Jena, Germany\\
  \small\textsuperscript{2}Competence Center Digital Research (zedif), Friedrich Schiller University Jena, Germany\\
  \small\textsuperscript{3}Computational Linguistics, Department of Linguistics, Bielefeld University, Germany\\
\\
  \small{
    \textbf{Correspondence:} \href{mailto:sarah.bachinger@uni-jena.de}{sarah.bachinger@uni-jena.de}
  }
}

\NewDocumentCommand{\otherfont}{v}{%
{\fontfamily{cmtt}\selectfont #1}%
}

\begin{document}
\maketitle
\begin{abstract}
  We apply NER to a particular sub-genre of legal texts in German:
  the genre of legal norms regulating administrative processes in public service administration.
  The analysis of such texts involves identifying stretches of text
  that instantiate one of ten classes identified by public service administration professionals. 
  We investigate and compare three methods for performing Named Entity Recognition (NER) to detect these classes:
  a Rule-based system, deep discriminative models, and a deep generative model.
  Our results show that Deep Discriminative models outperform both
  the Rule-based system as well as the Deep Generative model,
  the latter two roughly performing equally well, 
  outperforming each other in different classes.
  The main cause for this somewhat surprising result
  is arguably the fact that the classes used in the analysis
  are semantically and syntactically heterogeneous,
  in contrast to the classes used in more standard NER tasks.
  Deep Discriminative models appear to be better equipped for dealing with this heterogenerity
  than both generic LLMs and human linguists designing rule-based NER systems. 
  
\end{abstract}


\section{Introduction}

The application of Natural Language Processing (NLP) to legal texts in German is a relatively new development,
starting in an era where deep discriminative approaches to Named Entity Recognition (NER)
have already been established as state of the art technologies.
Hence, many implementations of NER for legal documents in German
turn to deep discriminative ML approaches directly, 
without systematically comparing these technologies to alternative approaches
\cite{leitnerFineGrainedNamedEntity2019,darjiGermanBERTModel2023,peikert2022extracting}

We aim to to fill this gap by comparing three different approaches: 
rule-based methods (symbolic AI), deep discriminative models and deep generative models. 
Moreover,
we run this comparison on a dataset that is very close to real-world applications
that NER in the legal domain may be used for, 
rather than re-using the standard NER benchmark datasets.
The application scenario for the GerPS-NER dataset 
that we chose to work with \citep{feddoul2024gerps} is to assist humans in analyzing legal bases\footnote{In Germany, the Federal Information Management \url{https://fimportal.de/glossar} 
provides standardized methods for analyzing such legal bases.} with the goal of creating digitized 
versions of administrative process schemata in the public administration.\footnote{Eventually, it is planned to integrate 
one or multiple approaches into a software for use by interested public administrations, 
so practical considerations regarding the approaches must also be kept in mind. }
Since the dataset we chose includes some highly structured sub-types of legal language,
we believe that it makes sense to include a rule-based approach in the comparison,
as this approach is known to be well suited for such texts. 

Each of the approaches we compare
comes with different trade-offs in terms of development effort and adaptability,
amounts of training data needed and prediction accuracy. 
While rule-based approaches can deal well with structured text, 
it is a time-consuming task to create the rulesets, they are relatively sensitive to errors in the dataset and can only 
detect known patterns. Deep generative systems bring with them a lot of contextual knowledge about
the world that the data is embedded in. On the other hand, they require large amounts of 
computational power, and they are expecting continuous text as input and output, making
them more difficult to work with when strict formats have to be adhered to.
Deep discriminative models have long been used in NLP tasks due to the relatively reliable
performance they deliver. The challenge with deep discriminative models is that they require
large amounts of training data.
\todo{probably need some references here too}


The remainder of this paper is organized as follows: 
\autoref{sec:relwork} outlines the relevant literature. 
\autoref{sec:concept} describes the methodology used in the study.
\autoref{sec:impl} describes the details of the model implementations.
\autoref{sec:results} presents the results, and \autoref{sec:disc} discusses their implications. 
Finally, \autoref{sec:conclusion} concludes the paper and suggests areas for future research.

\section{Related work}
\label{sec:relwork}

\subsection{NER in general}
\label{sec:ner-general}

General surveys of approaches to NER can be found in \citeauthor{yadavSurveyRecentAdvances2019} (\citeyear{yadavSurveyRecentAdvances2019})
and \citeauthor{pakhaleComprehensiveOverviewNamed2023} (\citeyear{pakhaleComprehensiveOverviewNamed2023}).
Both surveys discuss deep discriminative approaches,
which have become state of the art in NER tasks until recently,
when large language model-based approaches (in the following deep generative approach) appeared on the scene
\cite{wangGPTNERNamedEntity2023,bogdanovNuNEREntityRecognition2024,yeLLMDADataAugmentation2024,monajatipoorLLMsBiomedicineStudy2024,jungLLMBasedBiological2024,zhangLinkNERLinkingLocal2024,naguibFewShotClinical2024}. 
Different authors compare the effectiveness of LLM-based NER in the legal domain: for LegalLens,~\citeauthor{bernsohn-etal-2024-legallens} (\citeyear{bernsohn-etal-2024-legallens}) compared BERT models with open source LLMs on the domain of legal violation identification. 
\citeauthor{joshi-etal-2024-il} (\citeyear{joshi-etal-2024-il}) proposed ``IL-TUR, a benchmark for Indian Legal Text Understanding and Reasoning'' and offer among other things a LLM-based pipeline for the benchmark. 
With LAiW, \citeauthor{dai2023laiw} (\citeyear{dai2023laiw}) propose a benchmark for Chinese legal LLMs. \citeauthor{Bachinger2024} (\citeyear{Bachinger2024}) systematically evaluate different open source LLMs for their effectiveness in German text generation and use Prompt Engineering and Fewshot Prompting for NER on German legal texts. 
Their investigation on a small subset show optimistic results for one of their prompting schemes in combination with the German open source LLM LeoLM. 

While the success of the deep discriminative, 
and to a lesser degree the deep generative, approaches to NER
seems to have made rule-based approaches obsolete,
the simplicity and robustness of rule-based approaches
make them still strong competitors at least in some domains. 
For instance, \citeauthor{gorinskiNamedEntityRecognition2019} (\citeyear{gorinskiNamedEntityRecognition2019}) systematically compare
a rule-based NER system for electronic health records
with deep learning and transfer learning systems. 
They found that the hand crafted rule-based system
consistently outperforms both
the transfer learning and the deep learning systems,
reaching an overall F1-score of $0.95$.

In systematically comparing rule-based approaches
not only to deep discriminative,
but also to deep generative approaches
we hope to provide a broader evaluations
of the options currently available for NER applications in the legal domain.

\subsection{NER in legal documents}
\label{sec:ner-legal-documents}

NER systems for legal documents have been developed for a long time 
\cite{dozierNamedEntityRecognition2010}. 
Rule-based approaches to NER in legal documents
are mostly developed for languages lacking robust resources for ML development,
such as Afan Oromo \cite{m.e.phd.professordepartmentofcomputerscienceambouniversityamboethiopiaNLPRuleBased2019}
or Arabic \cite{abdallahIntegratingRuleBasedSystem2012}. 
The latter work stands out
by combining a rule-based approach with machine learning
and evaluating the effectiveness of both approaches.
The authors find
that the combined approach improves the F1-score
by $8-14\%$ compared to either the rule-based appraoch
or the machine learning approach alone.

The entities recognized by NER systems for legal documents
usually center around entities related to the court system
(\emph{judge, lawyer, court, court decision, jurisdiction, etc.})
and references to sections in law texts.
Our work seeks to find entities related
to legal norms for the administration of public services. 

\subsection{NER in German legal documents}
\label{sec:ner-german-legal}

NER systems for legal documents in German are mostly based on deep discriminative approaches. 
Thus \citeauthor{Leitner_2020} (\citeyear{Leitner_2020}) present a relatively large newly created 
dataset for German legal NER (German LER dataset) and also evaluate the performance 
of multiple differently configured BiLSTM-CRF models on this dataset.
\citeauthor{darjiGermanBERTModel2023} (\citeyear{darjiGermanBERTModel2023}) fine-tuned a German BERT model on
the German LER dataset and 
present their results that are better than the results originally achieved 
by \citeauthor{Leitner_2020} (\citeyear{Leitner_2020}) when presenting the dataset.
\citeauthor{Zollner_2021} (\citeyear{Zollner_2021}) used, among others, the German LER 
dataset when they compared the 
effect of different pre-training techniques for small BERT models and presented 
modified fine-tuning processes which resulted in performance improvements.
\citeauthor{erd2022evaluation} (\citeyear{erd2022evaluation}) used the same two architectures that will also 
be used in this paper (BiLSTM-CRF, XLM-RoBERTa \cite{Conneau_2019}) to evaluate and compare the 
performance improvements that different data augmentation methods and their 
combinations might achieve for NER tasks in the German legal domain.
\citeauthor{feddoul2024gerps} (\citeyear{feddoul2024gerps}) present GerPS-NER, a new corpus for NER on German legal texts
covering the sub-genre of legal norms regulating the administration of public services. 
In GerPS-NER, ten classes relevant for the analysis of this particular sub-genre are defined, 
which are intended to be used in aiding the digitization of public administration.
While the classes used by \citeauthor{Leitner_2020} (\citeyear{Leitner_2020}) resemble more common 
Named Entities, the GerPS-NER corpus also includes more abstract 
concepts, such as \emph{Bedingung} `condition' (see \autoref{sec:categ-annot-their}).
\autoref{sec:tokens-spans-example} illustrates 
how these classes are brought to bear on the analysis of legal norms. 

\section{Concept}
\label{sec:concept}
\begin{figure*}
    \includegraphics[width=\textwidth]{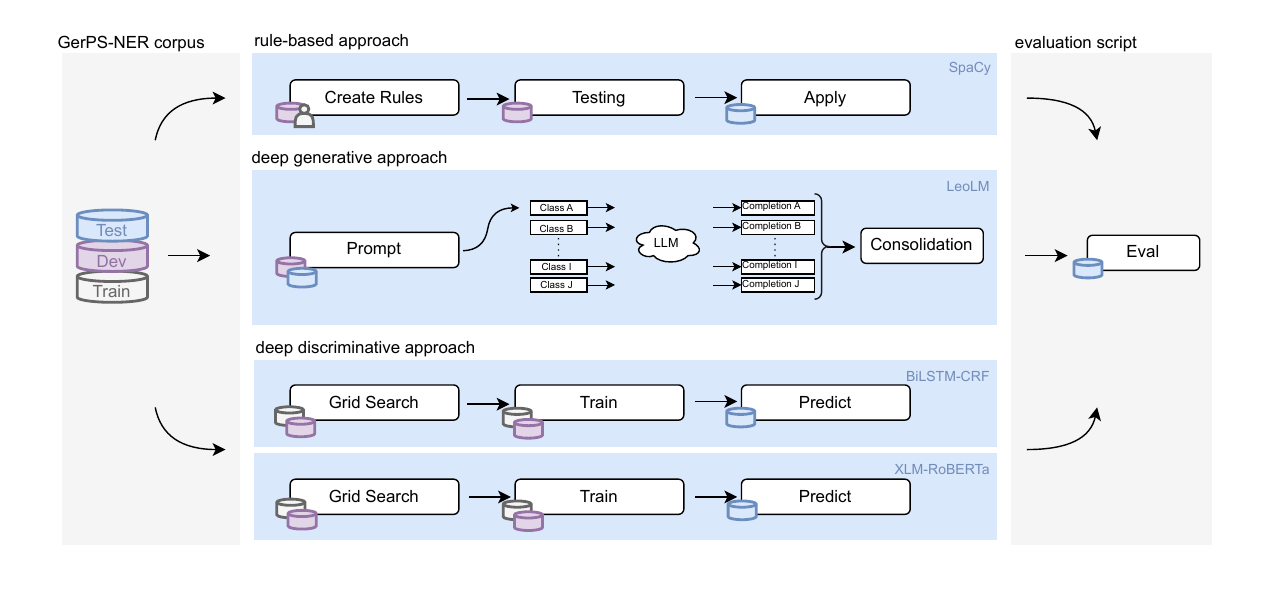}
    \caption{Overview of our workflow for comparing multiple machine learning approaches}\label{fig:concept}
   \end{figure*} 
The workflow for our experiments is shown in \autoref{fig:concept}.

\todo{We ought to list the classes and their definitions, perhaps in an appendix}
We use three different approaches to annotate German legal texts
for the occurance of expressions belonging to one of ten classes, as seen in \autoref{fig:concept}.
The classes were previously derived by GerPS-NER \cite{feddoul2024gerps}.

The first approach is a \textbf{rule-based approach} with a linguist drafting suitable
rules from gold standard examples.

To implement the \textbf{deep discriminative} approach, we selected two popular models that have also been used by \citeauthor{erd2022evaluation} (\citeyear{erd2022evaluation}).
The first is a BiLSTM-CRF model implemented with the \otherfont{FLAIR} framwork \cite{Akbik_2019_flair}, the second is the XLM-RoBERTa (XLM-R) \cite{Conneau_2019} transformer model,
implemented using the \otherfont{FLERT} extension \cite{Schweter_2020_flert} of the framework.


The \textbf{deep generative} approach is based on 
\citeauthor{Bachinger2024} (\citeyear{Bachinger2024}) who compared several LLMs (Large Language Models) for their performance on the task of legal norm analysis on a small dataset. We use the prompting scheme they deemed the best consisting of the task description, three examples per class and the annotation guideline. We also use the German LLM LeoLM~\cite{pluster2023leolm} for prompting to see whether the promising results for the small dataset hold up in a systematic evaluation.

The \textbf{dataset} used in the evaluation is GerPS-NER published by~\citeauthor{feddoul2024gerps} (\citeyear{feddoul2024gerps}). We adapted it slightly for our purposes as we found tokenization problems in the corpus. We split the corpus in 20\% development, 20\% test and 60\% training data. 
\todo{discuss this, tokenization issues (inconsistencies?) or just a different tokenizer? (spacy) or both?}
As shown in \autoref{fig:concept}, the train data split was only used by the deep discriminative approach, while the others used the dev split for creating rules and testing the code.


\section{Implementation}
\label{sec:impl}

In the following, we describe implementation details for the approaches. The code is available on zenodo~\cite{GerPS_zenodo}.
\subsection{Metrics for evaluation}
\label{sec:metrics}
As it is custom in the evaluation of NER tasks, we use the F1-score and the associated precision and recall values.
For the overall evaluation we use the macro F1-score since in our application-scenarion all classes are of equal importance.
However, our main focus is on the per-class scores. 

Measuring precision and recall,
and hence calculating the F1-score,
is essentially a token-based procedure.
However,
most of the entities we try to find
cover spans of several tokens (see \autoref{sec:tokens-spans-example}).
This opens up the possibility
that a prediction based on a certain rule
may not cover exactly the same span of tokens
as the ground truth. 
Token-based evaluation metrics
would systematically count spurious false positives
and false negatives and lead to lower values
of precision, recall and F1-score values
when prediction and ground truth only partially overlap. 
For these reasons we have decided
to supplement these token-based measures
with span-based ones.

The span-based measure we propose to use
is the intersection over union measure,
or Jaccard similarity score. 
This metric is commonly used in image processing tasks,
but \citeauthor{soleimaniNLQuADNonFactoidLong2021} (\citeyear{soleimaniNLQuADNonFactoidLong2021}) uses it for measuring text-span overlap
(text-span similarity).
We calculate the Jaccard score for each class individually.
Moreover,
two of our approaches,
the rule-based approach and the deep generative approach,
split the corpus in a large number of small files
containing one sentence each.
This means that we must aggregate the results for each class per file
and find a basis for a global evaluation.
To do this,
we determine the arithmetic mean of the Jaccard score values of all file inputs.
In addition,
we collect the number of inputs with Jaccard score 0,
i.e. cases of zero-overlap between prediction and ground truth,
and the number of total sentences where a given entity occurs.
This information helps to interpret the arithmetic mean of Jaccard scores.
After all,
relative low values of the mean of the Jaccard scores could be due to two different scenarios:
first,
there might be a certain number of sentences with a high Jaccard score
(near perfect overlap between prediction and ground truth),
and second,
all sentences may show some overlap between prediction and ground truth,
but not a lot.
Knowing the number of zero-overlap instances therefore
helps to shed light on the interpretation of the mean value.
Furthermore,
we can calculate the proportion of zero-overlap instances to the total number of instances
by dividing the former number with the latter,
yielding a number between 0 (best case) and 1 (worst case),
which furthermore sharpens the evaluation.
Finally,
we determine the median of the Jaccard scores.
Thus,
we base our evaluation of the respective approaches
on a variety of metrics that need to be carefully interpreted. 

\subsection{Rule-based approach}
\label{sec:rule-based-approach}

The \textbf{rulebased approach} is implemented in the \otherfont{SpaCy} framework, using token patterns and phrase patterns within \otherfont{SpaCy}'s \otherfont{EntityRuler}. 
Token patterns allow the use of morphological features in the rules.
Phrase patterns match against exact word or phrase matches
and are well suited to implement word-list based patterns (gazetteers).
Several phrase patterns are dynamically constructed programmatically by looping over the tokens of the input text, 
applying filter functions defined in Python.

The design choice to utilize \otherfont{SpaCy}'s EntityRuler helps to keep the rules simple and easily maintainable.
However, it also means that it is not possible to use syntactic information such as provided by \otherfont{SpaCy}'s dependency parser.
This in turn means that some patterns, for instance those defining the entity \emph{Bedingung} `condition', are inherently limited in the depths of recursion they can cover.

Patterns for the entities \emph{Handlungsgrundlage} `legal or formal grounds for the action described'
and \emph{Hauptakteur} `main actor' utilise word-list patterns (gazetteers) derived from the language model based on legal texts in German contained in the \otherfont{FLAIR} framework \otherfont{flair/ner-german-legal}.
These patterns are augmented by patterns derived from the development set.

\subsection{Deep discriminative approach}
\todo{verify stuff said here against code, and old data augmentation paper}

\todo{shorten? there is potential}
For the BiLSTM-CRF model, we use stacked German fastText \cite{Bojanowski_2017} and German forward and backward
\otherfont{FLAIR} embeddings \cite{Akbik_2018}. The model is trained using the default Stochastic Gradient Descent without 
momentum, with gradients clipped at $5$. The maximum number of training epochs is set 
to $150$, but the learning rate is annealed based on performance on the development set, with training stopping early 
if the learning rate drops below $0.0001$. Variational dropout is applied.

\todo{shorten? there is potential}
For the XLM-R model there are fewer parameters to configure. We chose this 
transformer model because preliminary studies on an early version of a subset of the corpus showed 
that it outperforms other German models we tested. The default setting in \otherfont{FLERT} is to 
use the \otherfont{AdamW} optimizer. We fine-tune the model with a 
learning rate that increases from $0$ to $5$ e-$6$ during the warm-up phase and then decreases linearly 
to $0$ by the end of the training.

\todo{shorten? there is potential}
For both models, we conducted a grid search to select the hyperparameters.
For the BiLSTM-CRF model, we found that a learning rate of $0.05$ (from the options of $0.05$, $0.1$, $0.2$) and a 
batch size of $16$ (from the options of $8$, $16$, $32$) produced the best results on the development set. 
For the XLM-R transformer model, $30$ fine-tuning epochs (from the options of $15$, $20$, $30$) combined with 
a batch size of $1$ (from the options of $1$, $4$) yielded the best performance.

\subsection{Deep generative approach}
\label{sec:dga}
The approach works as follows:
for every sentence from the dataset, ten prompts (one for each class) are created.
These prompts contain the class definition in addition to the components mentioned above.
The prompts are given to LeoLM and the completion is saved to a text file.
The completions are checked for their length and for the content of the predictions,
so that only predicted sentences containing the same tokens as the input sentence are processed further.
Next, the valid predictions are consolidated into one sentence.
Because there may be multiple predictions for a given token,
\citeauthor{Bachinger2024} (\citeyear{Bachinger2024}) use two different variants of sentence consolidation,
a so-called optimistic and a pessimistic sentence consolidation.
Optimistic sentences consolidation means that
if one of the model's possible predictions matches the ground truth,
this particular prediction is chosen.
Pessimistic sentence consolidation means that
if there are multiple predictions for a token, 
a new class X is assigned that represents conflicting annotations.
For each file in the test, we generate an optimistic (GenAI opt), a pessimistic (GenAI pes), and a gold standard IOB file from the dataset as the tokenization in this approach varies slightly from the dataset.



\section{Results}
\label{sec:results}
This section presents the results of the model evaluation, divided into two parts: 
an overview of overall performance and a detailed analysis of individual model performances. 

\subsection{Summary of key findings}
\todo{consistency: model vs approach}

\begin{figure*}
  \includegraphics[width=\textwidth]{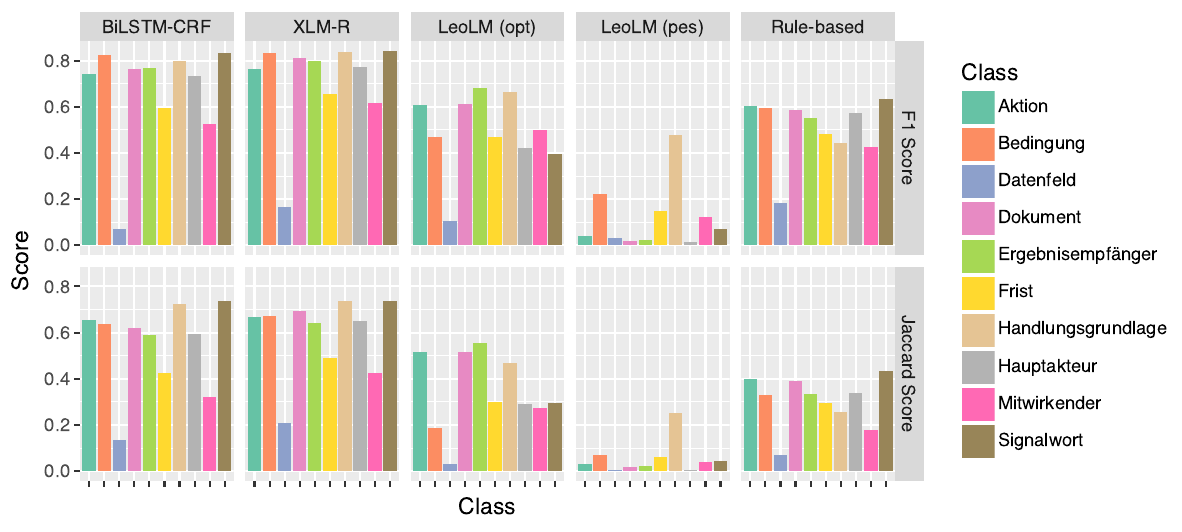}
  \caption{Evaluation results for the different approaches by class and score type.}
  \label{fig:plot}
 \end{figure*}

In this section, we outline our key findings. The micro F1-scores for the 
model predictions are presented in \autoref{tab:model_comparison_f1}, while the Jaccard 
score are shown in \autoref{tab:model_comparison_jacc}. Additionally, these results are visualized in 
\autoref{fig:plot}. Overall, the XLM-R model outperformed all other models across most classes,
except for the \emph{Datenfeld} `data field` class. Contrary to expectations, the deep generative 
model did not outperform the other models. Even the optimistic 
interpretation of its outputs resulted in the second-lowest macro F1-score, just above the 
pessimistic interpretation. The rule-based approach was the only one to surpass
the XLM-R model performance in at least one class and also outperformed 
the deep generative approach in terms of overall macro F1-score.

\begin{table*}[h!]
  \centering
  \fontsize{10pt}{12pt}\selectfont
  \begin{tabular}{c||c||c||c||c||c}
  Class & BiLSTM-CRF & XLM-R & LeoLM (opt) & LeoLM (pes) & Rule-based\\
  \hline
  Action & 0.7443 & \textbf{0.7621} & 0.6102 & 0.0421 & 0.6049\\
  Condition & 0.8244 & \textbf{0.8329} & 0.4678 & 0.2240 & 0.5944\\
  Data field & 0.0721 & 0.1676 & 0.1076 & 0.0338 & \textbf{0.1829}\\
  Document & 0.7661 & \textbf{0.8126} & 0.6144 & 0.0178 & 0.5861\\
  Recipient of service & 0.7674 & \textbf{0.8004} & 0.6828 & 0.0220 & 0.5531\\
  Deadline & 0.5967 & \textbf{0.6569} & 0.4699 & 0.1485 & 0.4813\\
  Legal grounds for action & 0.7985 & \textbf{0.8362} & 0.6643 & 0.4794 & 0.4450\\
  Main actor & 0.7315 & \textbf{0.7724} & 0.4239 & 0.0129 & 0.5747\\
  Contributor & 0.5276 & \textbf{0.6173} & 0.5020 & 0.1227 & 0.4258\\
  Signaling word & 0.8352 & \textbf{0.8423} & 0.3940 & 0.0701 & 0.6341\\
  \hline
  Macro F1-score & 0.6058 & \textbf{0.6455} & 0.4488 & 0.1067 & 0.5082\\

  \hline
  \end{tabular}
  \caption{F1-Scores for the evaluated approaches by class and model. The best score for each class is highlighted in bold.}
  \label{tab:model_comparison_f1}
  \end{table*}

  \begin{table*}[h!]
    \resizebox{\textwidth}{!}{%
    \centering
    \fontsize{10pt}{12pt}\selectfont
    \begin{tabular}{c||c|c||c|c||c|c||c|c||c|c}
    \multicolumn{1}{c||}{} & \multicolumn{2}{c||}{BiLSTM-CRF} & \multicolumn{2}{c||}{XLM-R} & \multicolumn{2}{c||}{LeoLM (opt)} & \multicolumn{2}{c||}{LeoLM (pes)} & \multicolumn{2}{c}{Rule-based} \\ 
    \hline
    Class &  mean ↑ & ratio ↓ & mean ↑ & ratio ↓ & mean ↑ &ratio ↓ &mean ↑ & ratio ↓ & mean ↑ & ratio ↓ \\
    \hline
    Action & 0.65 & 0.17 & \textbf{0.67} & 0.18 & 0.52 & 0.37 & 0.03 & 0.94 & 0.40 & 0.35 \\
    Contributor & 0.32 & 0.60 & \textbf{0.42} & 0.51 & 0.27 & 0.63 & 0.04 & 0.90 & 0.18 & 0.74 \\
    Main actor & 0.59 & 0.31 & \textbf{0.65} & 0.29 & 0.29 & 0.63 & 0.01 & 0.99 & 0.34 & 0.59 \\
    Recipient of service & 0.59 & 0.32 & \textbf{0.64} & 0.29 & 0.56 & 0.38 & 0.02 & 0.98 & 0.33 & 0.58 \\
    Deadline & 0.43 & 0.47 & \textbf{0.49} & 0.38 & 0.30 & 0.50 & 0.06 & 0.83 & 0.29 & 0.56 \\
    Condition & 0.64 & 0.29 & \textbf{0.67} & 0.26 & 0.19 & 0.54 & 0.07 & 0.70 & 0.33 & 0.51 \\
    Document & 0.62 & 0.29 & \textbf{0.69} & 0.22 & 0.52 & 0.43 & 0.02 & 0.98 & 0.39 & 0.50 \\
    Data field & 0.14 & 0.77 & \textbf{0.21} & 0.71 & 0.03 & 0.95 & 0.01 & 0.98 & 0.07 & 0.85 \\
    Signaling word & 0.74 & 0.19 & \textbf{0.74} & 0.19 & 0.29 & 0.63 & 0.05 & 0.93 & 0.44 & 0.49 \\
    Legal grounds for action & 0.72 & 0.21 & \textbf{0.74} & 0.20 & 0.47 & 0.35 & 0.25 & 0.52 & 0.26 & 0.35 \\
    \end{tabular}
    }
    \caption{Jaccard means for the evaluated approaches by class and model. The best score for each class is highlighted in bold.}
    \label{tab:model_comparison_jacc}
    \end{table*}

\subsection{Detailed performance analysis}
The following two sections will take a closer look at 
the individual model performances.

\textbf{F1-score}
\todo{reference to section in implementation}
Analyzing the results for the rule-based approach, the F1-scores for the respective classes 
are generally not very high, but range from $0.43$ to $0.63$. 
This indicates that there is not a lot of variation
in the performance of the rules implementing the various classes.
A similar trend is observed for the deep discriminative models, 
which perform consistently across all classes with a slightly 
broder and higher range, from $0.52$ to $0.84$.
The deep generative approach provides two evaluation reports (see \autoref{sec:dga}), 
with the F1-scores for the optimistic sentence consolidation exceeding those of the pessimistic one, as expected.
Notably, the \emph{Handlungsgrundlage} `legal grounds for action' class is less 
affected by the pessimistic consolidation scheme and remains the best-performing class by a 
significant margin. 
The \emph{Datenfeld} `data field' class is an exception for all mentioned models, with scores 
as low as $0.07$, $0.17$, $0.03$, $0.11$ and $0.18$ for BiLSTM-CRF, XLM-R, 
GenAI opt, GenAI pes and the rule-based approach, respectively,
with the rule-based approach achieving the highest score. The XLM-R model 
achieved twice the score of the BiLSTM-CRF for the \emph{Datenfeld} `data field` class, 
despite both models generally yielding similar results. 
The \emph{Datenfeld} `data field' class has proven to be notriously difficult to define,
annotate manually and capture in rules. 
It should therefore be considered an outlier and may best be excluded from consideration. 


\textbf{Jaccard}
\todo{add ref label to metrics section}
The Jaccard similarity score for a given class
provides a measure of how closely the text spans
marked as instantiating the class overlap
between prediction and gold standard,
in a given document.
Since the corpus is split into many documents
covering a sentence each,
we take a score for every sentence
and must look at the arithmetic mean value of these scores
in order to understand how well the system's prediction
for a given class performs.
However,
the mean of the Jaccard similarity scores
needs to be interpreted in context of
the zero-overlap to total count ratio
and the median value,
as discussed in \autoref{sec:metrics}. 
These values together provide another perspective on the model performance.
Overall, the mean Jaccard scores generally reflect the performance distribution 
across classes observed with the F1-score,
although they do give a somewhat different insight into the performance
of individual classes.
We discuss an example of the insights one can gain
from a close analysis of the Jaccard score
in the context of the Rule-based system in \autoref{sec:interpr-results-rule}. 
One thing to note is that, while the F1-scores and Jaccard means differ 
between the XLM-R and BiLSTM-CRF models, the ratios for the \emph{Datenfeld} `data field` class are quite similar,
at $0.77$ and $0.71$.
\todo{need more observatios here!}


\section{Discussion}
\label{sec:disc}




\subsection{Rule-based approach}
\label{sec:interpr-results-rule}

The token-based F-score evaluation requires little comment.
We therefore focus here on the span-based evaluation based
on the Jaccard similarity score (or Intersection-over-Union measure). 
In order to see the value of adding the Jaccard score analysis to the evaluation,
let us consider the performance of the classes \emph{Signalwort} `signaling word'
\emph{Handlungsgrundlage} `legal grounds for action' and \emph{Bedingung} `condition'.
We leave it to the reader to apply similar considerations
to the data from other classes in different approaches as listed in \autoref{tab:model_comparison_jacc}. 

Let us first consider the class \emph{Signalwort} `signaling word`.
This is the class with the highest Jaccard mean value,
a value of $0.43$.
The ratio of complete prediction failures (zero-overlap cases)
to the total number of occurances of the class
at $428:880$ is $0.49$.
In other words, 
there is a medium high number of zero-overlap cases.
This in turn means
that the implementation misses a significant number of instances of the class,
but when it does find an instance,
the span it marks as belonging to the class overlaps significantly with the gold standard.
This conclusion is reinforced by a relatively low median value of $0.32$.

Consider now the class \emph{Handlungsgrundlage} `legal grounds for action'.
This class as the second lowest ratio of zero-overlap cases ($216$) to total counts ($622$),
at a value of $0.35$.
The Jaccard score mean is $0.26$ ($0.258$).
This means that
the implementation successfully predicts significantly more test spans
belonging to this class than it misses,
but the overlap is not that large in many instances.
Again,
this conclusion is strenghtened by the fact that the median value of Jaccard scores is $0.25$.



It is interesting to note
that \emph{Handlungsgrundlage} `legal grounds for action' has the highest Precision value in the token-based evaluation at $0.78$
and \emph{Signalwort} `signaling word' the third highest at $0.73$.
Therefore it appears
that there is a close relation between the Precision value
and the entity ranking given by the zero-total ratio.
However,
the picture is complicated by the fact
that the second highest Precision value of $0.75$ is attached to the class \emph{Bedingung} `condition'.
But this class's zero-total ratio of $0.51$ suggests
that the system correctly predicts only less than half of the instances of the class
while the level of overlap in each case isn't very high either,
as indicated by a relatively low mean value of $0.33$.
This suggests that this class's implementation is less-well performing
than the token-based evaluation suggests.
It further illustrates
that although the token-based F-score evaluation
and the span-based Jaccard similarity score evaluation
roughly point in the same direction,
the span-based evaluation allows for a finer-grained interpretation of the system's performance. 

\subsection{Deep discriminative models}

The scores show that the \emph{Datenfeld} `data field' class is the most difficult class to predict, by a large margin.
One possible explanation for this could be that it 
requires much prior knowledge about the actual process and context to be able 
to judge whether something classifies as \emph{Datenfeld} `data field' or not. 
This idea is also supported by the fact that the transformer model solves 
this task better than the BiLSTM-CRF model (it achieves approx. twice the score). Besides that, 
the classes \emph{Frist} `deadline' and \emph{Mitwirkender} `contributor' are probably difficult to predict for the models, 
because even humans struggle to differentiate between \emph{Hauptakteur} `main actor' and \emph{Mitwirkender} `contributor'
and \emph{Frist} `deadline' and \emph{Bedingung} `condition' in many cases. This has an additioal adverse effect: 
it raises the probability that the annotations of these edge-cases are inconsistent 
and thereby makes learning harder for these difficult cases.

For both models the classes \emph{Signalwort} `signaling word' and 
\emph{Handlungsgrundlage} `legal grounds for action' are among the best-performing.
In the latter case this is expected, 
given that instances of the class \emph{Handlungsgrundlage} `legal grounds for action'
are relatively easy to identify based on their structure (e.g. \emph{§ 44b Absatz 1 Satz }).
The class \emph{Signalwort} `signaling word' has a more heterogeneous definition
(see \autoref{sec:categ-annot-their})
which would suggest that classification is more difficult.
However,
this class has an easily identifiable core
in the form of modal verbs or \emph{zu-}infinitive.
Apparently,
this core is significant enough that recognition proves to be robust. 

Regarding the Jaccard Scores it interesting to see that \emph{Handlungsgrundlage} `legal grounds for action'
also ranks second place with the 
BiLSTM-CRF model here even though its F1-score only ranks 5th. 
This is presumably because the spans for \emph{Handlungsgrundlage} `legal grounds for action' are 
generally long. This results in significant token overlap with most 
reasonable predictions, even if the start and end points are slightly inaccurate.

\subsection{Deep generative models}
In comparison, we see lower values for F1-score for this approach as compared to the work 
from \citeauthor{Bachinger2024} (\citeyear{Bachinger2024}), though in the latter, micro F1-score was used as a measure as 
compared to macro F1-score here. The best class for both metrics 
is \emph{Ergebnisempfänger} `recipient of service'. The generative
approach scores better in general according to the token-based evaluation, which might be 
due to the fact that the annotation scheme in the prompt is token based.

\subsection{Comparison}
\label{sec:comparison}

Based on our results, deep discriminative models outperform both 
the rule-based approach and deep generative models, with the rule-based 
approach slightly outperforming deep generative models.
That deep discriminative models perform well in this task
is consistent with other findings in NER research
\cite{yadavSurveyRecentAdvances2019,pakhaleComprehensiveOverviewNamed2023}; 
but that both rule-based and deep generative approaches
only reach the modest scores that we report is surprising,
given results in other published research. 
\citeauthor{gorinskiNamedEntityRecognition2019} (\citeyear{gorinskiNamedEntityRecognition2019}), for instance,
compare a rule-based NER system
with deep learning and transfer learning systems
in the domain of public health records.
They found that the hand crafted rule-based system
consistently outperforms both
the transfer learning and the deep leraning systems,
reaching an overall F1-score of $0.95$.
\citeauthor{wangGPTNERNamedEntity2023} (\citeyear{wangGPTNERNamedEntity2023}) claim
that while unsophistacted applications of LLM to NER
perform inferior to supervised learning models,
their suggested way of adapting LLMs to the NER task
improves performances to state of the art baseline levels.
\todo{and did we use their suggested way?}

That both the rule-based approach and the deep generative approach
take performance hits at roughly the same scale
strongly suggests that there must be a common cause affecting both approaches,
rather than individual causes having to do with the implementation of each.\footnote{%
  This is not to say that there may not be issues with our implementation of these approaches.
  On the contrary, we are well aware of limitations in the implementation particularly of the
  rule-based Approach. However, such limitations would only selectively affect 
  one approach and can not explain similarities in outcomes accross these approaches. 
  }
The most obvious cause lies in the definition of the classes used in the task.
As is apparent from the definitions and examples in \autoref{sec:categ-annot-their},
the classes are heterogenous with respect to semantic and syntactic types,
amalgamating linguistic categorization and legal or administrative classification schemas.
\citeauthor{gorinskiNamedEntityRecognition2019} (\citeyear{gorinskiNamedEntityRecognition2019}), in contrast,
had linguists designing the classes
in close cooperation with domain experts to come up with linguistically motivated entity classes, 
and \citeauthor{wangGPTNERNamedEntity2023} (\citeyear{wangGPTNERNamedEntity2023}) default to the general linguistically motivated
entity definitions such as \emph{Location} or \emph{Organization}. 
It appears
that human linguists and general LLMs both have similar difficulties processing heterogeneous classes.
deep discriminative models, on the other hand,
are able to learn heterogenous class patterns much more easily. 


\section{Conclusion and future work}
\label{sec:conclusion}
In this paper, we compared three different approaches for supporting legal norm analysis on 
German legal texts. We find that the deep discriminative models performed best in 9 out of 10 classes.
For class \emph{Datenfeld} `data field`, the rule-based approached performed better but the class is in general not well predicted. 
Future work may explore the integration of these methods, with a promising direction being the combination of deep discriminative approaches and rule-based techniques.
Previous research has found this combination to be productive,
c.f. for instance the work of \citeauthor{abdallahIntegratingRuleBasedSystem2012} (\citeyear{abdallahIntegratingRuleBasedSystem2012}).

Another intriguing option is the combination of the rule-based approach with the Deep Generative approach.
While the latter did not perform as well on the larger dataset compared to the results reported by \citeauthor{Bachinger2024} (\citeyear{Bachinger2024}),
it stands to reason that the combination with the rule-based approach
might improve the performance of both.
Since neither the rule-based approach nor the Deep Generative approach
requires the retraining of data models,
this combination could potentially keep implementation and development costs in a real-world scenario low.
One way to combine these approaches would be
to automatically extract the n best matches of the rule-based system for a given class
and use these as examples in the prompt for querying an existing LLM.

\section{Limitations}

For the deep discriminative approach, the hyperparameter optimization of both models was limited by the available resources.
Without such limitations, the search space for the grid search could have been extended to find better hyperparameters.

For the deep generative approach, there were some limitations due to using a pre existing implementation. The examples in the prompt were predefined and from a smaller data set. They may not be representative of the classes in the overall corpus. Also, the texts used in the related work by \cite{Bachinger2024} were annotated by different annotators and a different annotation guideline than other parts of the corpus, while maintaining the same classes. 

The rule-based system, in part due to early design decisions,
can not make use of syntactic information
such as dependency relations or phrase structure.
Moreover,
overlapping annotations and multiple annotations can not be used
because we had to stick to the CoNLL 2002 annotation scheme,
which does not allow for multiple NER annotations.
We plan to address these limitations in the near future.

\bibliography{literature}

\appendix

\newpage

\section{GerPS-NER dataset classes}
\label{sec:categ-annot-their}
\todo{have appendix in header or not?}

\begin{enumerate}
\item[] \textbf{Hauptakteur} `main actor' -- The office or person that is mainly responsible
  for the administration of the service. E.g. \emph{Agentur für Arbeit} `Federal Employment Agency'
\item[] \textbf{Ergebnisempfänger} `recipient of service' -- Person or company applying
  to receive the benefits of the service in question. E.g. \emph{Antragsteller} `applicant'
\item[] \textbf{Mitwirkender} `contributor' -- External office or actor that needs to give input
  at specific points in the administration of the service. E.g. \emph{Deutsches Patent- und Markenamt}
  `German Patent and Trade Mark Office'
\item[] \textbf{Aktion} `action' -- Action carried out by one of the actors in the course
  of the administration of the service. E.g. \emph{erteilen} `to grant'
\item[] \textbf{Dokument} `document' -- Documents that the actors exchange between them.
  E.g. \emph{Antrag} `application form'
\item[] \textbf{Signalwort} `signaling word' -- Word or expression influencing the degree of obligatoriness
  of a decision made ont he basis of this statute. E.g. modal verbs such as \emph{kann} `may';
  \emph{zu-}Infinitiv \emph{Die Genehmigung ist zu erteilen} `permission is to be granted';
  adjectives or adverbs such as \emph{angemessen} `appropriate' or \emph{berechtigt} `being eligible';
  phrases such as \emph{auf Wunsch} `if desired' 
\item[] \textbf{Bedingung} `condition' -- Preconditions for taking an action.
  Mostly expressed by conditional clauses. 
\item[] \textbf{Frist} `deadline' -- Time limits for certain steps in the administrative process;
  temporal preconditions. E.g. \emph{spätestens am zehnten Tage vor der Wahl} `at the latest on the tenth day
  before the election'
\item[] \textbf{Datenfeld} `data field' -- Expressions that indicate the content of a data field in a form. 
  E.g. \emph{Vollständige Anschrift} `complete address'
\item[] \textbf{Handlungsgrundlage} `legal grounds for action' -- Cross reference to the legal basis
  for the administrative process in question. E.g. \emph{§3, Absatz 2 des Patentgesetzes}
  `Paragraph 3, section 2 of the patent law'
  \todo{fix linebreak, maybe by shortening Signalwort}
\end{enumerate}

\onecolumn
\section{F1-score evaluation results}
\label{tab:full_f1}

\begin{table*}[h!]
  \resizebox{\textwidth}{!}{
  \centering
  \begin{tabular}{c||ccc||ccc||ccc||ccc||ccc}
  Class & \multicolumn{3}{|c|}{BiLSTM-CRF} & \multicolumn{3}{|c|}{XLM-R} & \multicolumn{3}{|c|}{LeoLM (opt)} & \multicolumn{3}{|c|}{LeoLM (pes)} & \multicolumn{3}{|c|}{Rule-based} \\
  \hline
   & F1 & P & R & F1 & P & R & F1 & P & R & F1 & P & R & F1 & P & R \\
  \hline
  Action & 0.74 & 0.73 & 0.76 & 0.76 & 0.75 & 0.77 & 0.61 & 0.69 & 0.55 & 0.04 & 0.05 & 0.04 & 0.60 & 0.60 & 0.61 \\
  Condition & 0.82 & 0.80 & 0.85 & 0.83 & 0.85 & 0.82 & 0.47 & 0.57 & 0.40 & 0.22 & 0.29 & 0.18 & 0.59 & 0.75 & 0.49 \\
  Data field & 0.07 & 0.10 & 0.07 & 0.17 & 0.20 & 0.15 & 0.11 & 0.06 & 0.64 & 0.03 & 0.02 & 0.21 & 0.18 & 0.14 & 0.26 \\
  Document & 0.77 & 0.77 & 0.76 & 0.81 & 0.81 & 0.81 & 0.61 & 0.69 & 0.56 & 0.02 & 0.02 & 0.02 & 0.59 & 0.65 & 0.53 \\
  Recipient of service & 0.77 & 0.78 & 0.76 & 0.80 & 0.80 & 0.80 & 0.68 & 0.73 & 0.64 & 0.02 & 0.02 & 0.02 & 0.55 & 0.64 & 0.49 \\
  Deadline & 0.60 & 0.64 & 0.56 & 0.66 & 0.68 & 0.64 & 0.47 & 0.71 & 0.35 & 0.15 & 0.25 & 0.11 & 0.48 & 0.63 & 0.39 \\
  Legal grounds for action & 0.80 & 0.79 & 0.81 & 0.84 & 0.82 & 0.86 & 0.66 & 0.87 & 0.54 & 0.48 & 0.67 & 0.37 & 0.44 & 0.78 & 0.31 \\
  Main actor & 0.73 & 0.70 & 0.77 & 0.77 & 0.76 & 0.78 & 0.42 & 0.37 & 0.50 & 0.01 & 0.01 & 0.02 & 0.57 & 0.57 & 0.58 \\
  Contributor & 0.53 & 0.52 & 0.54 & 0.62 & 0.62 & 0.62 & 0.50 & 0.52 & 0.49 & 0.12 & 0.13 & 0.12 & 0.43 & 0.44 & 0.41 \\
  Signaling word & 0.84 & 0.81 & 0.86 & 0.84 & 0.83 & 0.86 & 0.39 & 0.46 & 0.34 & 0.07 & 0.08 & 0.06 & 0.63 & 0.73 & 0.56 \\
  \hline
  Micro Average & 0.79 & 0.77 & 0.81 & 0.81 & 0.81 & 0.81 & 0.51 & 0.59 & 0.45 & 0.17 & 0.17 & 0.17 & 0.56 & 0.69 & 0.47 \\
  \end{tabular}
  }
  \caption{F1, Precision and Recall scores for the evaluated approaches by class and model.}
  \label{tab:f1_full_results}
  \end{table*}

\onecolumn
\section{Tokens and spans in example annotation}
\label{sec:tokens-spans-example}

An example of the gold standard annotation of Corpus$/$corpus\_v2$/$test$/$1009.conll is given in \autoref{tab:example_annotation}.
Notice that class annotations typically span multiple tokens.
This is typically the case in classes that are mostly associated with linguistic expressions at the clause level,
such as \emph{Bedingung} `condition.'
But also classes which are often expressed by single token spans such as \emph{Signalwort} `signaling word'
(see token 8) 
can at times span multiple tokens,
as is the case in this example \emph{im Einvernehmen} `with approval' in tokens 26--27  
(indicating that the main actor is not completely free in the determination of the action
but must involve another agency as a contributor). 

\begin{table*}[h]
  \centering
  \begin{tabular}{l l l }
    Nr & Token & IOB-class \\ \hline
    1 & \emph{Das} `the' & O \\
    2 & \emph{Bundesamt} `federal office' & B-\emph{Hauptakteur} `main actor' \\
    3 & \emph{für} `for' & I-\emph{Hauptakteur} `main actor' \\
    4 & \emph{Sicherheit} `security' & I-\emph{Hauptakteur} `main actor' \\
    5 & \emph{in} `in' & I-\emph{Hauptakteur} `main actor' \\
    6 & \emph{der} `the' & I-\emph{Hauptakteur} `main actor' \\
    7 & \emph{Informationstechnik} `information technology' & I-\emph{Hauptakteur} `main actor' \\
    8 & \emph{kann} `may' & B-\emph{Signalwort} `signaling word' \\
    9 & \emph{bei}  `with' & B-\emph{Bedingung} `condition' \\
    10 & \emph{Mängeln} `shortcomings' & I-\emph{Bedingung} `condition' \\
    11 & \emph{in} `in' & I-\emph{Bedingung} `condition' \\
    12 & \emph{der} `the' & I-\emph{Bedingung} `condition' \\
    13 & \emph{Umsetzung}  `implementation' & I-\emph{Bedingung} `condition' \\
    14 & \emph{der} `of the' & I-\emph{Bedingung} `condition' \\
    15 & \emph{Anforderungen} `requirements' & I-\emph{Bedingung} `condition' \\
    16 & \emph{nach} `according to' & I-\emph{Bedingung} `condition' \\
    17 & \emph{Absatz} `paragraph' & I-\emph{Bedingung} `condition' \\
    18 & \emph{1d}  & I-\emph{Bedingung} `condition' \\
    19 & \emph{oder} `or' & O \\
    20 & \emph{in} `in' & B-\emph{Bedingung} `condition' \\
    21 & \emph{den} `the' & I-\emph{Bedingung} `condition' \\
    22 & \emph{Nachweisdokumenten} `proof certificates' & I-\emph{Bedingung} `condition' \\
    23 & \emph{nach} `according to' & I-\emph{Bedingung} `condition' \\
    24 & \emph{Satz} `sentence' & I-\emph{Bedingung} `condition' \\
    25 & \emph{1}  & I-\emph{Bedingung} `condition' \\
    26 & \emph{im} `with the' & B-\emph{Signalwort} `signaling word' \\
    27 & \emph{Einvernehmen} `approval' & I-\emph{Signalwort} `signaling word' \\
    28 & \emph{mit} `of' & O \\
    29 & \emph{der} `the' & O \\
    30 & \emph{Bundesnetzagentur} `federal network agency' & B-\emph{Mitwirkender} `contributor' \\
    31 & \emph{die} `the' & O \\
    32 & \emph{Beseitigung} `removal' & O \\
    33 & \emph{der} `of the' & O \\
    34 & \emph{Mängel} `shortcomings' & O \\
    35 & \emph{verlangen} `require' & B-\emph{Aktion} `action'\\
    36 & \emph{.} & O \\
    
  \end{tabular}
  \caption[Example annotation]{Annotation of the sentence `The federal office for security in information technology can in case of shortcomings against the requirements of paragraph 1d or in the proof certificates according to sentence 1 require, with the approval of the federal network agency, the correction of the shortcomings'. }
  \label{tab:example_annotation}
\end{table*}

\end{document}